%% file: main.tex
\documentclass[11pt, a4paper, logo]{googledeepmind}

\pdftrailerid{redacted}

\makeatletter
\renewcommand\bibentry[1]{\nocite{#1}{\frenchspacing\@nameuse{BR@r@#1\@extra@b@citeb}}}
\makeatother

\usepackage{kantlipsum, lipsum}
\usepackage{dsfont}
\usepackage{gdm-colors}

\usepackage[authoryear, sort&compress, round]{natbib}

\graphicspath{{figures/}}

\title{MusicRL: Aligning Music Generation to Human Preferences}

\correspondingauthor{gcideron@google.com, agostinelli@google.com}

\author{Geoffrey Cideron}
\author{Sertan Girgin}
\author{Mauro Verzetti}
\author{Damien Vincent}
\author{Matej Kastelic}
\author{Zalán Borsos}
\author{Brian McWilliams}
\author{Victor Ungureanu}
\author{Olivier Bachem}
\author{Olivier Pietquin\footnote{Now at Cohere}}
\author{Matthieu Geist\textcolor{blue}{$^1$}}
\author{Léonard Hussenot}
\author{\hspace{1cm}Neil Zeghidour\footnote{Now at Kyutai}}
\author{Andrea Agostinelli}
\affil{Google DeepMind}

\usepackage{subfigure}
\usepackage{natbib}

\newcommand{\musiclm}{{MusicLM}}
\newcommand{\musicrlhf}{\texttt{MusicRL}}
\newcommand{\musicrlhfu}{\texttt{MusicRL-U}}
\newcommand{\musicrlhfr}{\texttt{MusicRL-R}}
\newcommand{\musicrlhq}{\texttt{MusicRL-Quality}}
\newcommand{\musicrlhm}{\texttt{MusicRL-MuLan}}
\newcommand{\musicrlhfru}{\texttt{MusicRL-RU}}

\begin{abstract}
We propose \musicrlhf{}, the first music generation system finetuned from human feedback. Appreciation of text-to-music models is particularly subjective since the concept of musicality as well as the specific intention behind a caption are user-dependent (e.g. a caption such as ``upbeat workout music'' can map to a retro guitar solo or a technopop beat). Not only this makes supervised training of such models challenging, but it also calls for integrating continuous human feedback in their post-deployment finetuning. \musicrlhf{} is a pretrained autoregressive \musiclm{} \citep{musiclm} model of discrete audio tokens finetuned with reinforcement learning to maximize sequence-level rewards. We design reward functions related specifically to text-adherence and audio quality with the help from selected raters, and use those to finetune \musiclm{} into \musicrlhfr{}. We deploy \musiclm{} to users and collect a substantial dataset comprising 300,000 pairwise preferences. Using Reinforcement Learning from Human Feedback (RLHF), we train \musicrlhfu{}, the first text-to-music model that incorporates human feedback at scale. Human evaluations show that both \musicrlhfr{} and \musicrlhfu{} are preferred to the baseline. Ultimately, \musicrlhfru{} combines the two approaches and results in the best model according to human raters. Ablation studies shed light on the musical attributes influencing human preferences, indicating that text adherence and quality only account for a part of it. This underscores the prevalence of subjectivity in musical appreciation and calls for further involvement of human listeners in the finetuning of music generation models. Samples can be found at \href{https://google-research.github.io/seanet/musiclm/rlhf/}{google-research.github.io/seanet/musiclm/rlhf/}.

\end{abstract}

\begin{document}

\maketitle

\input{content}

\end{document}

%% file: content.tex
\section{Introduction}
\label{intro}

\begin{figure*}[ht]
\vskip 0.2in
\begin{center}
\centerline{\includegraphics[width=.9\columnwidth]{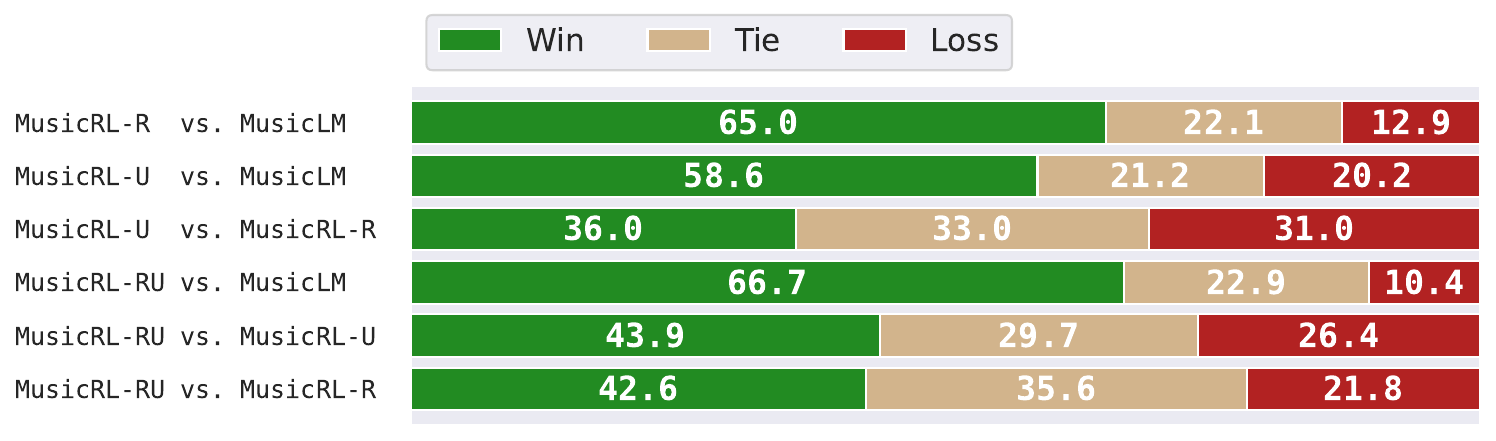}}
\caption{Results of the qualitative side-by-side evaluation for the RLHF finetuned models. In each X vs. Y comparison, the green bar corresponds to the percentage of times model X was preferred, the yellow bar to the percentage of ties and the red bar to the percentage of times model Y was preferred. \musicrlhfr{} is the \musiclm{} model finetuned on quality and text adherence reward. \musicrlhfu{} is finetuned on a reward model of user preferences. \musicrlhfru{} is finetuned sequentially on quality and adherence to text and then on a reward model of user preferences. While every RLHF finetuned version of \musiclm{} significantly outperforms \musiclm{}, \musicrlhfr{} and \musicrlhfu{} achieve comparable performance, while \musicrlhfru{} is overall the preferred model.}
\label{side-by-side-eval}
\end{center}
\vskip -0.2in
\end{figure*}

\looseness=-1
Generative modelling of music has experienced a leap forward: while it was until recently either limited to the fine modelling of individual instruments \citep{nsynth, sing, engel2020ddsp} or the coarse generation of polyphonic music \citep{jukebox}, models can now handle open-ended, high-fidelity text-controlled music generation \citep{riffusion,musiclm,audioldm,musicgen}. In particular, text-to-music systems such as MusicLM \citep{musiclm} and MusicGen \citep{musicgen} build on audio language models, as they cast the generative process as an autoregressive prediction task in the discrete representation space of a neural audio codec \citep{soundstream, defossez2022highfi}. While this approach has demonstrated its ability to generate realistic speech \citep{audiolm, vall-e, soundstorm}, sound events \citep{audiogen} and music, it suffers from a few shortcomings. First, the next-token prediction task used to train these systems --- while generic enough to model arbitrary audio signals --- lacks any prior knowledge about musicality that could bias those towards generating music that is more appealing to listeners. Second, while the temperature sampling used at inference allows for generating diverse audio from a single text caption, this diversity is only desirable along certain axes such as melody or performance, while musicality and adherence to the prompt should remain consistently high.

\looseness=-1
These fundamental issues of autoregressive generative models have been extensively observed and addressed in the context of language modelling. For example, several works have explored finetuning machine translation models to maximise the BLEU score \citep{ranzato_sequence_level, google_translate} or summarization models to improve the relevant ROUGE metric \citep{ranzato_sequence_level, summarization_rl, ferret_summarization}. Such metrics are typically sequence-level, and evaluate the output of a non-differentiable sampling process (e.g., greedy decoding, temperature sampling). This is typically circumvented by using a reinforcement learning method which models the metric of interest of a reward function and the generative model as a policy. The underlying algorithmic similarity between such text generation systems and autoregressive music models suggests that --- given the proper reward functions--- one could use reinforcement learning to improve music generation.

\looseness=-1
Music generated given a prompt should exhibit three properties: adherence to the input text, high acoustic quality (absence of artifacts), and ``musicality'' or general pleasantness. Automatic metrics have been proposed to quantify the text adherence like Classifier KLD \citep{yang2022diffsound} or MuLan Cycle Consistency \citep{musiclm} as well as acoustic quality with Fr\'echet Audio Distance \citep{fad}. Such metrics could be used as reward functions. Yet, designing automatic proxies to measure musicality is challenging. Most of the previous approaches \citep{jaques2017sequence,kotecha2018bach2bach,guimaraes2017objective,latif2023survey} rely on complex music theory rules, are restricted to specific musical domains (e.g., classical piano) and only partially align with human preferences. This gap between automatic metrics and human preferences has again been extensively studied in language modelling, with RLHF (Reinforcement Learning from Human Preferences) becoming the \textit{de facto} way of aligning conversational models \citep{gpt4, team2023gemini} with human feedback.

Human preferences as referred in previous work \citep{ouyang2022training, stiennon2020learning} mainly refers to the preferences of raters. Raters may not be representative of the population interacting with the model (e.g. rating services such as Amazon Mechanical Turk\footnote{https://www.mturk.com/} uses a global workforce). Especially in the context of music, this population gap can have a significant impact on the preferences \citep{crosscultural}. Collecting large scale user preferences data could help bridge the population gap by collecting considerably more interactions in contrast with raters.

\looseness=-1
In this work, we introduce \musicrlhf{}, a text-to-music generative model finetuned with reinforcement learning. Starting from a \musiclm{} baseline, we use an automatic measure of text adherence as well as a new acoustic fidelity metric as reward functions to perform RL finetuning. Human evaluations indicate that generations from the resulting \musicrlhfr{} are preferred over those from \musiclm{} 83\% of the time, as measured by $win/(win+loss)$.
Then, to explicitly align the model with human judgment, we collect a dataset of pairwise preferences from users interacting with \musiclm{} to fit a reward model. Ablation studies on the reward model trained on user interaction data demonstrate that user preferences strongly correlate with musicality. Extensive human evaluations reveal that the music generations coming from the resulting \musicrlhfu{} are preferred over the base model 74\% of the time. Finally, we combine automatic rewards and human feedback to finetune \musicrlhfr{} into \musicrlhfru{} and show that this model outperforms all alternatives more than 62\% of the time. To the best of our knowledge, this work is the first attempt at leveraging human feedback at scale to improve an audio generative model.

\section{Related Work}
\label{related-work}

\textbf{Music generation.} 
While earlier approches to musical audio generation were limited in terms of producing high quality outputs \citep{jukebox} or semantically consistent long audios \citep{perceiverAR}, recent research has achieved a level of quality that allows for an enjoyable listening experience. A first line of work casts the task of music generation as categorical prediction in the discrete token space provided by a neural audio codec~\citep{soundstream,defossez2022highfi}, and trains a Transformer-based~\citep{vaswani2017attention} model for next token prediction~\citep{audiolm} or parallel token decoding~\citep{soundstorm,vampnet,parker2024stemgen}. Combining this generative backbone with text-conditioning either through a text encoder or text-audio embeddings~\citep{elizalde2022clap,mulan} provides high-quality text-to-music models ~\citep{musiclm,musicgen}. A parallel line of work relies on diffusion models and casts the task of music generation as denoising of audio waveforms and spectrograms~\citep{Noise2music} or learned latent representations~\citep{schneider2023mousai, audioldm}. In 
both cases, the models are trained offline on a collection of existing musical recordings and inference is run in a stochastic fashion (e.g. diffusion or temperature sampling), which provides diversity but also uncertainty on the outputs (e.g. in terms of text-adherence or quality). Previous work~\citep{spear} has circumvented this issue by sampling many sequences, ranking them with a score function (e.g. a reference-free audio quality estimator) and returning the best candidate. This considerably increases inference cost and requires well-defined score functions.

\musicrlhf{} addresses these limitations by finetuning a MusicLM~\citep{musiclm} model with reinforcement learning, using reward functions derived from automatic metrics, small scale high-quality human ratings, and large scale user feedback. To the best of our knowledge, \musicrlhf{} is the first music generation system that shows the benefits from integrating feedback from hundreds of thousands of users.

\looseness=-1
\textbf{RL-finetuning of music generation models.} Most previous works in RL-finetuning music generation models involve designing handmade reward signals based on principles of music theory \citep{jaques2017sequence,kotecha2018bach2bach,guimaraes2017objective,latif2023survey} or simple patterns like repetitions \citep{squiggles}. \citet{jaques2017sequence} uses a set of rules inspired by a melodic composition theory \citep{gauldin1988practical} (e.g., stay in key, play motifs and repeat them, avoid excessively repeating notes) in combination with a KL regularization term. These approaches have several limitations: the rule sets can be incomplete or contradictory, practitioners must find the correct balance between different rewards, and the rules themselves derive from music theory, which is an imperfect approximation of human musical preferences. \citet{rlduet} finetune an online music accompaniment generation model with four reward models learned from data and rule-based reward that assign -1 when a note is excessively repeated. Each reward model corresponds to the probability of a chunk of the generation given a context (the context and the chunk to predict is different for each reward). These rewards are learned with a masked language model \citep{devlin-etal-2019-bert} loss on a music dataset. Yet, such methods only apply to restricted musical domains (e.g. monophonic piano) or symbolic generation.
In contrast with previous work,  \musicrlhf{} learns human preferences from its own raw audio generations. This allows for improving music generation across the whole spectrum of musical genres and styles, from lo-fi hip-hop to orchestral symphonies and modal jazz.

\looseness=-1
\textbf{RL from human feedback.} RLHF recently became a critical step in the training of conversational models used in applications such as Bard \citep{gemini2023} or GPT-4 \citep{openai2023gpt4}. RLHF has first been applied to solve Atari games \citep{christiano2017deep} before being used widely, for example in natural language tasks \citep{ziegler2019fine, stiennon2020learning, ouyang2022training, jaquesrlhf, bai2022training} or in image generation \citep{imagerlhf,wallace2023diffusion}. \citet{wallace2023diffusion} uses Direct Optimisation Algorithm (DPO) \citep{dpo} to finetune a diffusion model on human preference data. To the best of our knowledge, we are the first to apply RLHF to music generation models.

\begin{figure}[t]
\begin{center}
\centerline{\includegraphics[width=0.7\columnwidth]{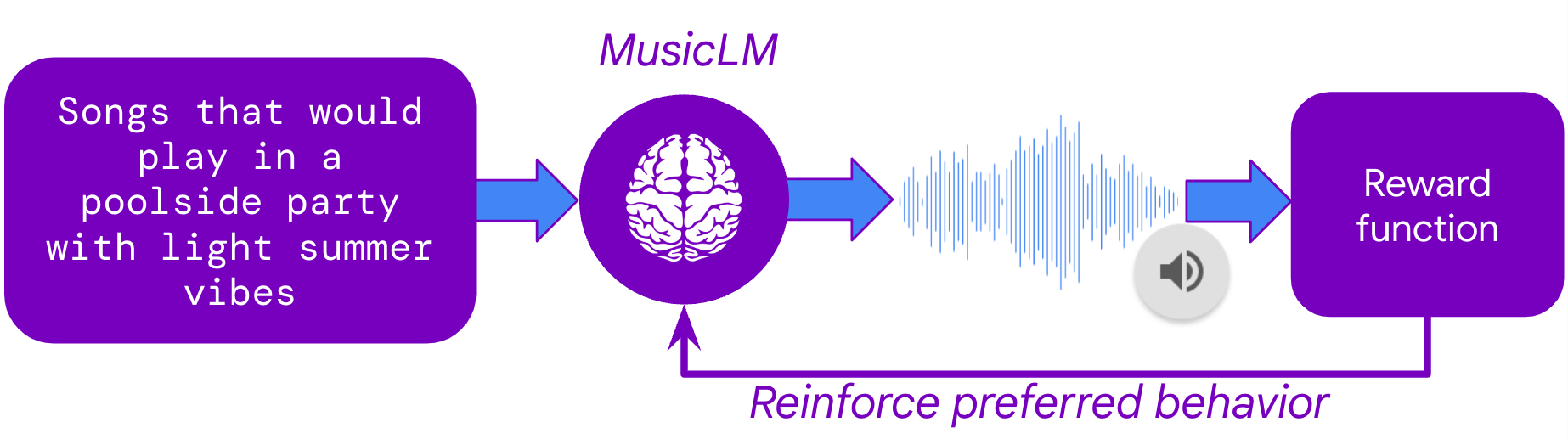}}
\caption{Given a dataset of music captions, \musiclm{} generates audio samples that are scored with a reward function. The RL algorithm finetune the model to maximise the received reward.}
\label{training-loop}
\end{center}
\end{figure}

\section{Method}

\subsection{MusicLM}
\label{method-musiclm}
\musiclm{}~\citep{musiclm} is an autoregressive model for generating music from text descriptions. Following the design of AudioLM~\citep{audiolm}, MusicLM relies on two different types of audio representations for generation: \emph{semantic} tokens, which are quantized representations of masked audio language models such as w2v-BERT~\citep{wav2vec-bert} and \emph{acoustic} tokens, the discrete representations produced by neural audio codecs such as SoundStream ~\citep{soundstream}. While the semantic tokens ensure the long-term structural coherence of the generation process, the acoustic tokens allow for high-quality synthesis. To ensure high-bitrate reconstructions, SoundStream uses residual vector quantization (RVQ) --- a stack of vector quantizers where each quantizer operates on the residual produced by the previous quantizers --- to discretize the continuous audio representations, imposing a hierarchical structure on the acoustic tokens. Additionally, \musiclm{} relies on MuLan~\citep{mulan}, a joint music-text contrastive model, for conditioning the audio generation task on descriptive text.

\looseness=-1
\musiclm{} was initially introduced as a 3-stage Transformer-based autoregressive model. The first stage learns the mapping between MuLan and semantic tokens. The second stage predicts the first levels from the output of the SoundStream RVQ (coarse acoustic tokens) from MuLan and semantic tokens. The last stage predicts the remaining SoundStream RVQ levels (fine acoustic tokens) from coarse acoustic tokens.

For the purpose of RL finetuning, we choose to optimise the semantic and coarse acoustic modelling stages, which are the most important contributors to acoustic quality, adherence to the text and overall appeal of the generated music. We address the challenges of jointly optimizing semantic and coarse acoustic modelling by using a single autoregressive stage that operates on frame-interleaved semantic and acoustic tokens. While simplifying the RL setup and problem formulation, this approach increases modeled token sequence length. We address this with a hierarchical transformer, similarly to ~\citet{rq-transformer, yu2023megabyte, yang2023uniaudio}.
Finally, instead of the original autoregressive fine acoustic modelling stage of \musiclm, we use Soundstorm~\citep{soundstorm} for achieving efficient parallel generation.

For simplicity, by referring to \musiclm{} in this work, we refer only to the autoregressive modelling stage of interleaved semantic and coarse acoustic tokens, which is the text conditioned modelling stage that can be finetuned with RL.

\subsection{RL finetuning procedure} 
\looseness=-1
We use the standard formulation of RL in the context of finetuning large language models as done in previous work \citep{ziegler2019fine}. Figure~\ref{training-loop} illustrates the RL training loop.
The agent acts according to its policy $\pi_\theta$ with $\theta$ the weights that parameterize the policy. The policy is an autoregressive model taking as input ${a_0,\dots,a_{t-1}}$, the sequence of previously generated tokens and outputs a probability distribution over the next action, i.e., the next token to pick : $a_t \sim \pi_\theta(.|a_0\dots a_{t-1})$. The RL finetuning phase aims at maximizing $\mathbb{E}_{\pi_{\theta}} [\sum_{t}r(a_0\dots a_t)]$ with $r$ a given reward function.
We use a KL regularized version of the REINFORCE algorithm \citep{pg, jaques2017sequence} to update the policy weights. Given a trajectory $(a_t)^T_{t=0}$ and denoting $s_t=(a_0... a_{t-1})$, the corresponding policy gradient objective to maximise is 
\vspace{-0.1cm}
\begin{equation*}
\begin{split}
\mathbb{J}({\theta}) = (1-\alpha) [&\sum_{t=0}^T \log \pi_\theta(a_t|s_t) (\sum_{i=t}^T r(s_i) - V_\phi(s_t))] \\
- \alpha &\sum_{t=0}^T \sum_{a \in A}[\log(\pi_\theta(a|s_t)/\pi_{\theta_0}(a|s_t))],
\end{split}
\end{equation*}

with $A$ the action space which here corresponds to the codebook, $\alpha$ the KL regularization strength, and $V_\phi$ the baseline. The baseline value function $V_\phi$ is used to decrease the variance in the policy gradient objective \citep{sutton2018reinforcement} and it is trained to estimate the mean return of the current policy. The baseline is learned as follows: $$\min_{\phi} \mathbb{E}_{\pi_{\theta}} \sum_t(\sum_{k=t}^Tr(s_k) - V_\phi(s_t))^2.$$
Both the policy and the value function are initialized from the initial \musiclm{} checkpoint with weight $\theta_0$.

\subsection{Reward Signals}
\label{reward-signals}

\textbf{Text adherence.} We derive a reward model for text adherence from pretrained MuLan~\citep{mulan} embeddings. MuLan is a contrastive audio-text embedding model trained on music clips and weakly-associated, free-form text annotations. We compute the cosine similarity between the text embedding of the input prompt and the audio embedding of the generated music, resulting in a reward value in $[-1;1]$. We refer to this metric as MuLan score. Because our models generate 30-second audios, while MuLan is trained on 10-second audio clips, we divide each audio into three segments, we calculate MuLan scores for each segment, and we average the results.

\looseness=-1
\textbf{Acoustic quality.} Another main attribute of musical generation is acoustic quality, e.g. whether a clip sounds like a professional recording or is contaminated with artifacts. We rely on a reference-free quality estimator trained to predict the human Mean Opinion Score (MOS - between 1 and 5) of a 20 second music clip. We train the model on a mix of human-created and \musiclm{}-generated music clips, where each clip was rated by 3 raters. The raters were tasked to judge only the acoustic quality, to avoid confounding factors such as musicality. We refer to this metric as the quality score. Because our models generate 30-second clips, we compute quality scores on the first 20 seconds and on the last 20 seconds, and average the two scores.

\begin{figure}[t]
\vskip 0.2in
\begin{center}
\centerline{\includegraphics[width=1.0\columnwidth]{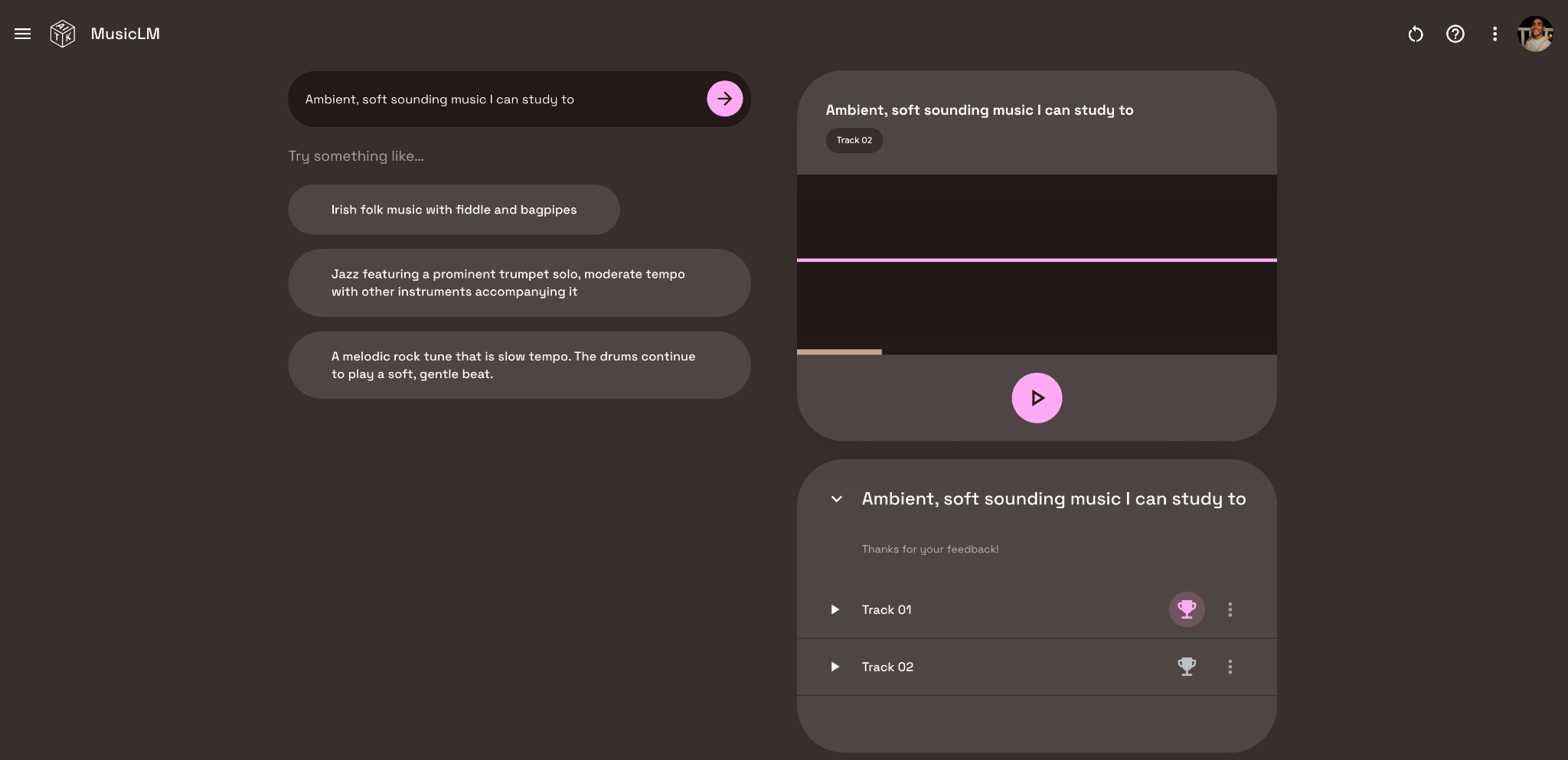}}
\caption{The AI Test Kitchen MusicLM interface. The user can write a prompt or choose from suggestions. Each prompt generates two 20s clips, and the user can label their favorite clip among the two with a trophy.}
\label{aitk-platform}
\end{center}
\vskip -0.2in
\end{figure}

\looseness=-1
\textbf{User preferences.}
\label{user-preference-data}
We deploy the pretrained text-to-music \musiclm{} model through the AITK web-based interface \footnote{\url{https://aitestkitchen.withgoogle.com/}} to a large scale userbase. We choose to collect feedback through pairwise comparisons~\citep{christiano2017deep}: when a user seizes a prompt, we generate two 20s candidate clips and let the user optionally assign a trophy to one of them. An important design choice implied by this process is the absence of specific instructions, which is intended not to bias users towards precise musical attributes and rather communicate their overall subjective taste. We only consider preferences from users that listen to both generations. After filtering, we obtain a dataset of pairwise user data of size 300,000. This dataset minimizes the biases that often arise from human raters (as detailed in Appendix~\ref{r-to-u}).%

\looseness=-1
Our reward model takes as input the caption's text and corresponding audio tokens and outputs a scalar score. This model is trained with a Bradley-Terry Model~\citep{bradley1952rank} as in \citet{christiano2017deep}, which enables learning a pointwise ELO score from pairwise preferences. It is initialized with the \musiclm{} checkpoint, as first results demonstrated that, starting from scratch, the reward model was not able to do better than chance at predicting human preferences.
We split the user preference dataset into a train split of size 285,000 and an evaluation split of size 15,000. After training for 10,000 steps on batches of 32 pairs, the reward model achieves 60\% of accuracy on the evaluation set (see Figure~\ref{rm-input-length-and-no-text}).

\looseness=-1
To pre-assess the performance of the reward model, we conduct an internal small-scale human evaluation on 156 audio comparisons from the user preference dataset. In 60\% of cases, our team's preferences aligned with the established preferences in the dataset. This result is comparable to the performance of the reward model. Furthermore, this low agreement rate highlights the inherent subjectivity in judging music preferences, compared to domains such as summarization where \citet{stiennon2020learning} estimated at 73-77\% the agreement rate for the OpenAI human preference dataset.
When finetuning \musiclm{} on the user preference reward model, since our models generate 30-second audios, we average the scores computed from the first and last 20 seconds of audio.

\section{Experimental Setup}
\label{experimental-setup}

\subsection{Datasets}
Given the pretrained reward signals as described in Section~\ref{reward-signals}, the RL finetuning step uses a dataset exclusively composed of captions, used for prompting all \musiclm{}-based models. Consequently, no ground-truth audio is involved in the finetuning process.
We follow the same procedure as \citet{Noise2music} for synthetically generating captions from three sources. We use the LaMDA model~\citep{thoppilan2022lamda} to generate descriptions of 150,000 popular songs. After providing song titles and artists, LaMDA's responses are processed into 4 million descriptive sentences about the music. We split 10,028 captions from MusicCaps~\citep{musiclm} into 35,333 single sentences describing music. Furthermore, we collect 23,906 short-form music tags from MusicCaps.
Additionally, we extend the previous captions with the 300,000 prompts collected from users, as described in Section~\ref{user-preference-data}.
We randomly split the data, using 90\% for training and 10\% for evaluation.

\subsection{Training procedure}
\looseness=-1
In the following experiments, we RL-finetune the \musiclm{} model with the same RL algorithm and the same hyperparameters. The common decoding scheme is temperature sampling with temperature $T=0.99$. The temperature was chosen with subjective inspection to have a good quality-diversity tradeoff for the generations of  \musiclm{}. The RL-finetuned models differs only with the reward function employed during their training process.

\looseness=-1
\textbf{\musicrlhfr{}.} We RL-finetune \musiclm{} for 20,000 training steps (1) with the MuLan reward, (2) with the quality reward, and (3) with a linear combination of the MuLan and the quality reward: the resulting models are respectively called  \musicrlhm{}, \musicrlhq{}, and \musicrlhfr{}.
Throughout our experiments, we normalize the quality reward from $[1;5]$ to $[0;1]$ as preliminary experiments have shown that the combination of the MuLan and the quality reward gives the best results when both rewards are on the same scale. We still display in figures the un-normalized scores.

\textbf{\musicrlhfu{}.} We RL-finetune \musiclm{} for 5000 training steps with the user preference reward model to obtain a model that we call \musicrlhfu{}.

\textbf{\musicrlhfru{}.} To combine all the reward signals, we RL-finetune \musicrlhfr{} for 1000 training steps on the user preference reward model. For this experiment, the KL regularization is computed between the model being finetuned and \musicrlhfr{}. The resulting model is called \musicrlhfru{}. We find that the sequential approach of first finetuning on MuLan and quality and then finetuning on the user preference reward outperforms learning from the three rewards at the same time. We hypothesize this comes from the fact that it takes a small number of gradient steps (under 2000) before over optimizing on the user preference reward while it takes around 10,000 steps to optimise the other rewards. Moreover, using the user preference reward model in a final stage in this matter may allow the model to align better on the human preferences.

\subsection{Evaluation}
\looseness=-1
The main metrics we report in our experiments are the quality reward, the MuLan reward, and the user preference reward model. We report the metrics either against the training step to show progress along the training, or against the KL divergence to the base model. This is typically used as a proxy to measure the distance to the base checkpoint and thus the retention of the original capabilities of the model~\citep{christiano2017deep, ferret_summarization}.

For the qualitative evaluation, we use 101 diverse, internally-collected prompts, representing a balanced range of musical genres (see Appendix~\ref{qualitative-evaluation} for the full list). We use these prompts to generate audio samples from each evaluated model. We select raters for their experience listening to varied musical styles ($>$6 years) and fluency in written English. During the qualitative evaluation, raters are presented with two audio clips generated by different models using the same text prompt. We ask raters to rate each clip on a scale of 1 to 5, considering adherence to the text prompt, acoustic quality and overall appeal to the audio clip.
Each comparison is performed by three different raters, totaling 303 ratings per model comparison. From these ratings, we compute a win rate metric which is defined as $win/(win+loss)$.

\looseness=-1
\subsection{Checkpoint selection}
For all RL-finetuned models, we select the best checkpoint by inspecting the quantitative results and listening to the music generations. For \musicrlhfr{}, \musicrlhfu{}, and \musicrlhfru{} we respectively choose the checkpoint after 10,000 training steps, 2000 training steps, and 1000 training steps.

\section{Results}
\label{results}

We aim to answer the following questions: (1) Can RL-finetuning on MuLan and quality rewards improve the generation quality of text-to-music models such as \musiclm{}? (2) Can RLHF improve the alignment of the generated music to generic preferences from users? (3) Is it possible to combine all reward signals to further improve performance?

\subsection{Quantitative Results}
\label{quant-results}

\begin{figure}[ht!]
\centering
\begin{subfigure}{}
    \includegraphics[width=.48\textwidth]{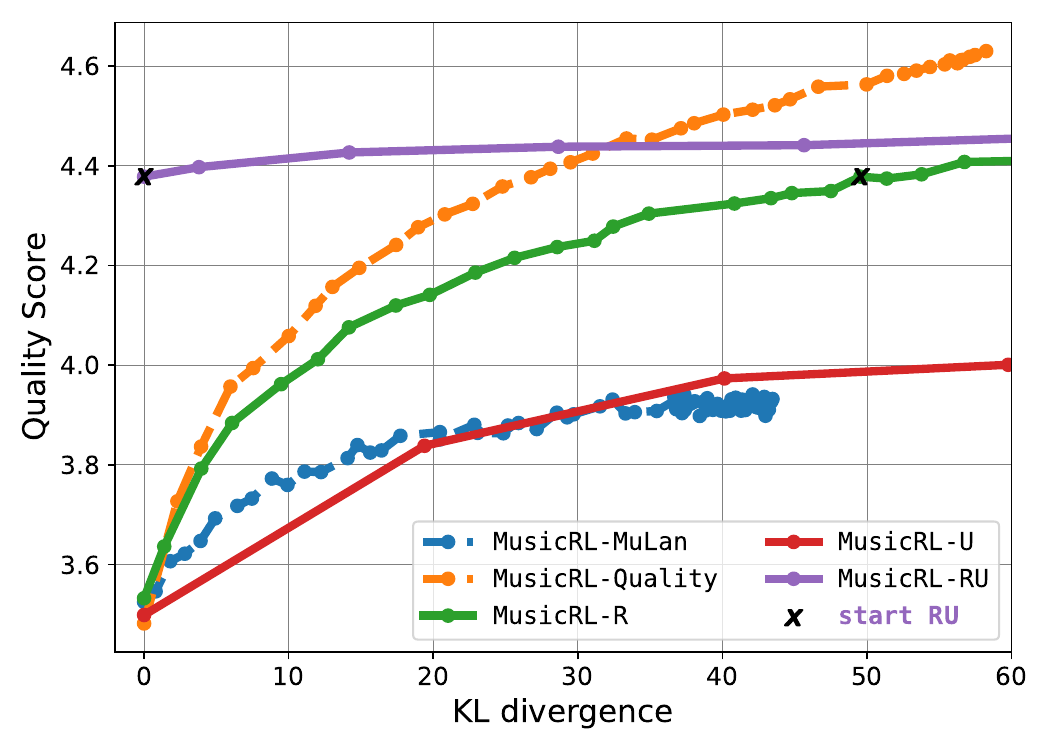}
    \label{mosic-vs-kl}
\end{subfigure}
\hfill
\begin{subfigure}{}
    \includegraphics[width=.48\textwidth]{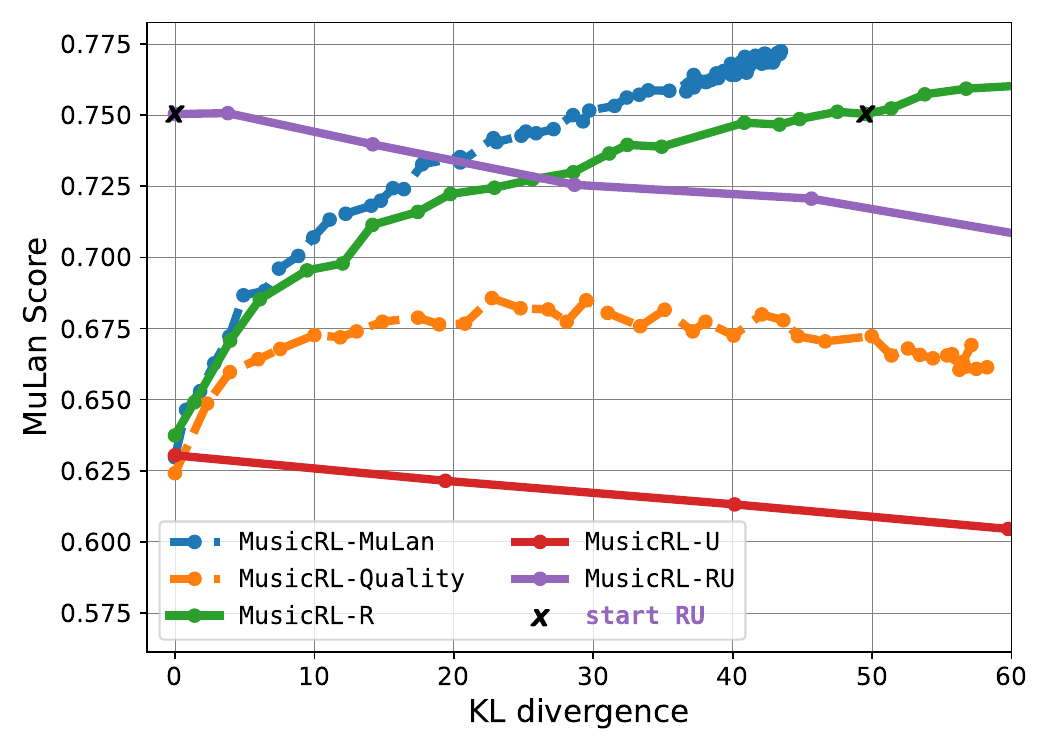}
    \label{mulan-vs-kl}
\end{subfigure}
\hfill
\caption{Quality (left) or MuLan score (right) vs KL divergence for the RL-finetuned models.  The KL divergence is computed between the RL-finetuned models and \musiclm{} except for \musicrlhfru{} where the KL divergence is computed against \musicrlhfr{}. The black cross corresponds to the checkpoint used to start the training of \musicrlhfru{}. RL-finetuning successfully optimises the quality and the MuLan scores (\musicrlhfr{}). Additionally, optimizing the user preference reward (\musicrlhfru{}, \musicrlhfru{}) improves the quality score while marginally decreasing the MuLan score.}
\label{mosic-mulan-vs-kl}
\end{figure}

\begin{figure}[ht!]
\centering
\includegraphics[width=0.48\textwidth]{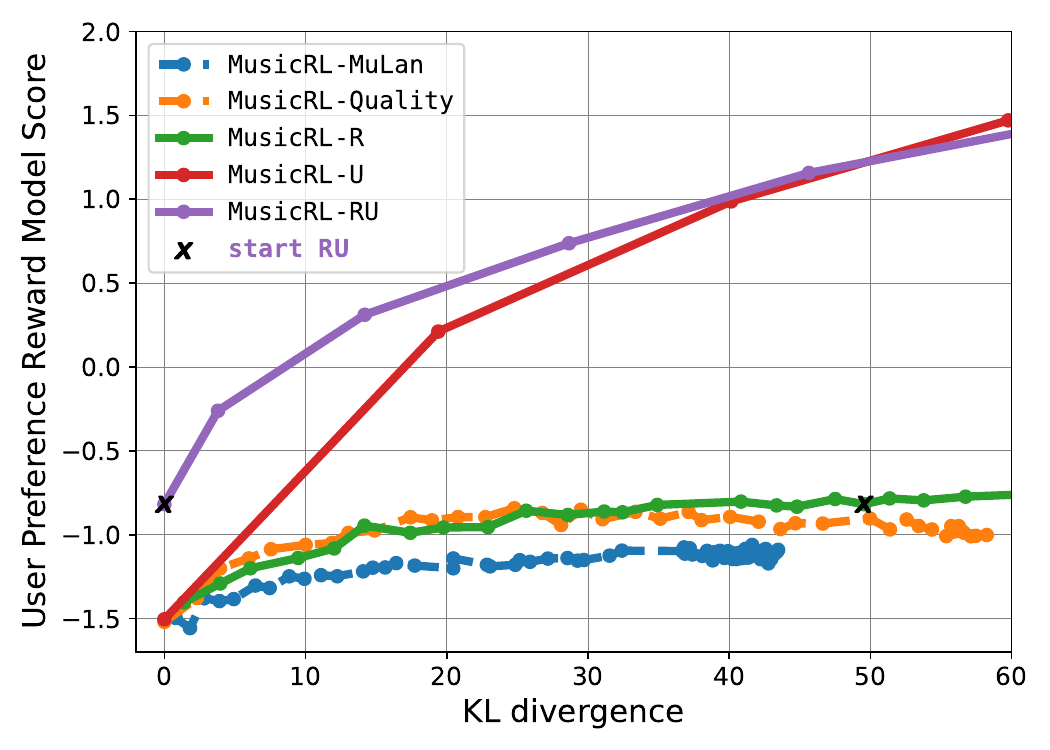}
\caption{User Preference Reward Model Score for the different RL-finetuned models. The KL divergence is computed between the RL-finetuned models and \musiclm{} except for \musicrlhfru{} where the KL divergence is computed against \musicrlhfr{}. The black cross corresponds to the checkpoint used to start the training of \musicrlhfru{}. RL-finetuning successfully improves the user preference reward model score of the generations (see \musicrlhfu{} and \musicrlhfru{} curves). When trained on other rewards (MuLan and/or quality) the user preference reward model score slightly improves.}
\label{fig:rm-vs-kl}
\end{figure}

In all quantitative evaluations, we analyse model progress during RL finetuning by tracking scores of rewards against the KL divergence from the initial model. Regardless of whether we train with a single reward model or a combination of both as in \musicrlhfr{}, we evaluate model performance on all reward signals.

\looseness=-1
Figure~\ref{mosic-mulan-vs-kl} shows that RL-finetuning successfully optimises both quality and MuLan scores. Specifically, finetuning on the quality reward alone leads to the greatest increase in quality score (from 3.5 MOS to 4.6 MOS), and a smaller increase in the MuLan score (from 0.58 to 0.61). Conversely, finetuning on only the MuLan reward maximises the MuLan score (from 0.58 to 0.71), with a less pronounced quality score improvement (from 3.5 MOS to 4.1 MOS). Leveraging both quality and MuLan rewards significantly improves both scores (quality: 3.5 MOS to 4.4 MOS; MuLan: 0.58 to 0.71), while marginally increasing KL divergence. Given the promising and stable performance in simultaneously optimizing MuLan and quality scores, we perform qualitative evaluations only on \musicrlhfr{}.

\looseness=-1
Figure~\ref{rm-vs-step} (in Appendix~\ref{curves-vs-step}) shows that after 10,000 finetuning steps on the quality reward, the reward model trained on user preference begins assigning lower scores to music samples. This suggests that finetuning solely on the quality reward is prone to reward over-optimisation \citep{coste2023reward,rame2024warm,rlduet}.

Figure~\ref{fig:rm-vs-kl} demonstrates that finetuning with the user preference reward model significantly improves generation scores, increasing them from -1.5 to over 1.5. Figure~\ref{mosic-mulan-vs-kl} shows that despite not training on the quality reward, the quality score increases from 3.5 MOS to 4 MOS. The MuLan score slightly decreases from 0.58 to 0.55. Yet, Figure~\ref{mosic-mulan-vs-step} highlights that over-optimizing the user preference reward model can drastically reduce the MuLan score. Overall, this suggests that user preference feedback particularly enhances audio quality while having minimal impact on text adherence.

Figure~\ref{fig:rm-vs-kl} shows that optimizing the user preference reward model on a model finetuned for 10,000 steps on quality and MuLan improves the user preference reward model score significantly. Figure~\ref{mosic-mulan-vs-kl} shows that the quality score slightly increases while the MuLan score slightly decreases, which confirms the impact of the user preference reward model observed in the previous paragraph.

\subsection{Qualitative Results}

\begin{table}
\begin{center}
\begin{tabular}{ l r r }
\toprule
 Model & MOS & \# wins \\ 
\midrule
 \musiclm{} & 3.07 & 133 \\  
 \musicrlhfr{} & 3.54 & 362 \\    
 \musicrlhfu{} & 3.54 & 372 \\   
 \musicrlhfru{} & 3.82 & 460 \\ 
\bottomrule
\end{tabular}
\end{center}
\caption{Average mean opinion score (MOS) and number of wins across all rating tasks, for each model. The music generated from the RL-finetuned models are significantly scored higher in average than the ones from \musiclm{}. The best performing model both in term of MOS and number of wins is \musicrlhfru{}.}
\label{mos-win-table}
\end{table}

Figure~\ref{side-by-side-eval} presents human rater evaluations of pairwise comparisons between all possible model combinations across \musiclm{}, \musicrlhfr{}, \musicrlhfu{} and \musicrlhfru{}.
When compared to \musiclm{}, \musicrlhfr{} wins 65\% of the time, ties 22.1\% and loses 12.9\%. This translates into a 83\% win rate in favor of \musicrlhfr{}.
\musicrlhfu{} is also strongly preferred over \musiclm{} as it achieves a 74\% win rate against \musiclm{}.
\looseness=-1
The best performing model overall is \musicrlhfru{}. When compared to \musiclm{}, \musicrlhfru{} is strongly preferred by the raters with a 87\% win rate. When compared to the other RL-finetuned models, \musicrlhfru{} achieves a win rate of 66\% against \musicrlhfr{}, and 62\% against \musicrlhfu{}. All results described above are statistically significant according to a post-hoc analysis using the Wilcoxon signed-rank test \citep{DBLP:reference/stat/ReyN11}.
 
Table~\ref{mos-win-table} summarises results from all qualitative evaluations by showing average mean opinion score (MOS) and number of wins across all rating tasks, for each model. On both metrics, all RL-finetuned models outperform \musiclm{}, with \musicrlhfru{} being the best performing model.

Lastly, \musicrlhfr{} and \musicrlhfu{} perform comparably according to raters, as shown in Figure~\ref{side-by-side-eval} and Table~\ref{mos-win-table}.

\subsection{Takeaway}
Our results demonstrate several key findings: (1) \musicrlhfr{} shows that RL-finetuning on text adherence and quality rewards improves the generation quality of \musiclm{}; (2) \musicrlhfu{} confirms the ability to leverage generic user preferences data to improve \musiclm{}; (3) \musicrlhfru{} outperforms all other models, demonstrating that the above reward signals are complementary and can be combined for the highest performance.

\section{Understanding Human Feedback Through the Lens of the Reward Model}
\label{rm-analysis}

\begin{figure}[t]
\begin{center}
\centerline{\includegraphics[width=.5\columnwidth]{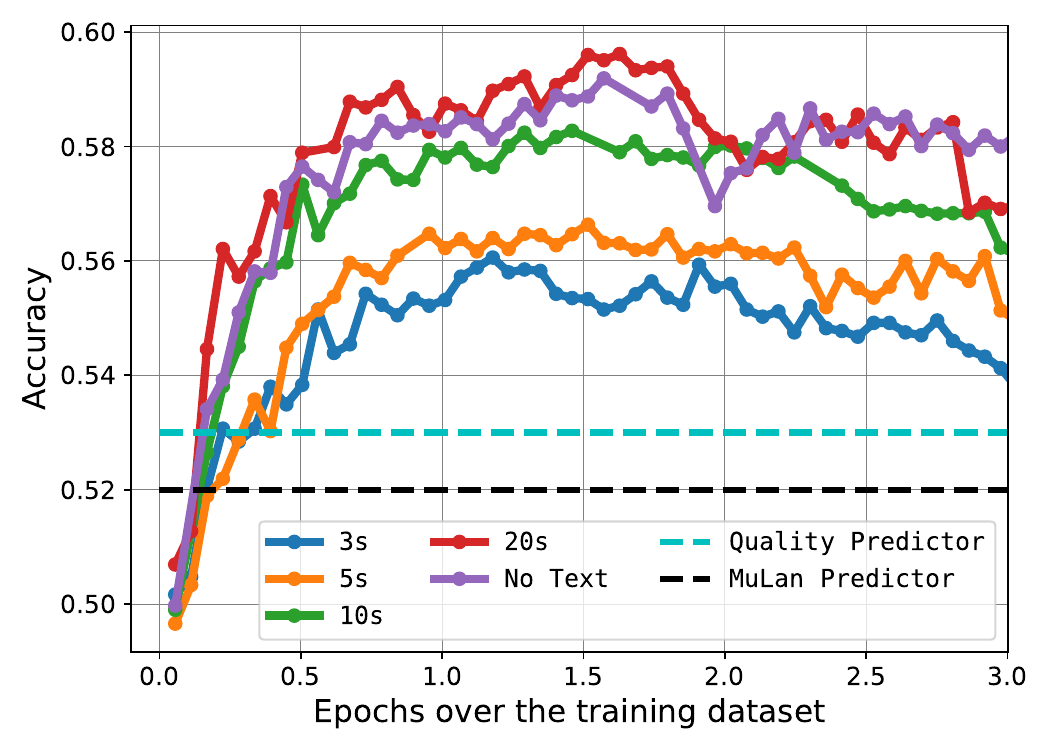}}
\caption{Ablations on the user preference reward model. The reward model is learned either with no text tokens (No Text) or with a cropped version of the input audio (i.e. 10s, 5s, 3s). While dropping the text tokens does not significantly impact the accuracy of the reward model, cropping the audio substantially degrades performance. This suggests that text adherence and audio quality are not the primary factors influencing user audio preferences, as additionally shown by the low accuracy when using text adherence based (MuLan) or audio quality based predictors for user preference.}
\label{rm-input-length-and-no-text}
\end{center}
\vskip -0.3in
\end{figure}
In this section, we analyse reward model accuracy to uncover the specific music elements that influence user preferences. This analysis directly addresses our research question: What is the user paying attention to when rating the audio?

We categorise generated music into three components which might drive the users' choice on their audio preferences: (1) text adherence, (2) audio quality, and (3) musicality. In particular, defining and modelling musicality is a complex task, which underscores our focus on human feedback as a solution, moving beyond rule-based limitations.

\looseness=-1
\subsection{Importance of the text input}
\label{subsec:text_importance}
To isolate the impact of text on pairwise preference evaluation, we drop text tokens while training the reward model. Accuracy remains stable as shown by Figure~\ref{rm-input-length-and-no-text}.
Additionally, we measure how often the music clip with the highest MuLan score corresponds to the preferred one. On the evaluation set, these indicators only match 51.6\% of the time, which is very close to random accuracy.
Overall, these findings indicate that adherence to the text prompt was not a primary driver of human preference in our experiment.
This aligns with our quantitative results in Section~\ref{quant-results}, which show no significant improvement in text adherence as measured by MuLan, when training \musicrlhfu{}.

\subsection{Importance of the audio quality} Since audio quality remains relatively consistent within a generated sequence, a few seconds of audio should provide sufficient information to evaluate this aspect. We train reward models on different input audio tokens length corresponding to 10, 5, and 3 seconds. As shown in Figure~\ref{rm-input-length-and-no-text} the evaluation accuracy on pairwise preference decreases as we reduce the length of the input tokens, dropping from 60 to 56\% when using 3-5 seconds of input audio.
The significant accuracy decrease suggests that other musical components play a complementary role in user preference.
Additionally, we replicate the analysis done in~\ref{subsec:text_importance} and measure how often the music clip with the highest quality score is preferred. As shown in Figure~\ref{rm-input-length-and-no-text} the quality predictor achieves 53.3\% accuracy on the evaluation dataset.
These findings indicate that audio quality is not the only driver of human preference, while being a better signal than text adherence.
This is consistent with our quantitative results in Section~\ref{quant-results}, where training \musicrlhfu{} improves marginally on the quality score.
Overall, this analysis shows that user preference is influenced by music elements which go beyond text adherence and audio quality.

\section{Limitations and Future Work}

\textbf{Aligning feedback and evaluation.}
When training on user preference data, a limitation of our current setup is the \textit{population gap} between those who provide feedback to improve the model (general users) and those who assess the results (selected raters). A direction for future work is to directly measure the perceived improvements from the user's perspective.

\textbf{Using on-policy data.}
For the reasons explained in Section~\ref{method-musiclm}, in this work we collected user preferences on a different version of \musiclm{} compared to the one used for RL finetuning. A clear path for improvement is to iteratively collect on-policy data (data generated by the model that is being finetuned) and use it to update the model. Eventually, this would allow for real integrated feedback where finetuned models are continuously deployed to collect new feedback while improving the user experience.

\textbf{Refining the user preference dataset.}
Several interesting research directions involve refining the large user interaction dataset. For instance, identifying and retaining examples where users express a confident and clear preference could reduce noise and improve the overall dataset quality. Furthermore, focusing on techniques to train robust reward models on smaller, but highly relevant datasets could facilitate research directions such as model personalization for specific users.

\section{Conclusion}
\looseness=-1
In this work, we introduce \musicrlhf, the first text-to-music generative model aligned with human preferences. In a first set of experiments, we derive sequence-level reward functions that inform on the adherence to the caption as well as the acoustic quality. When finetuning a pretrained \musiclm{} model to optimise these rewards with RL, the quantitative and qualitative results show consistent improvements over the pretrained baseline. We then show for the first time that we can align music generation with generic preferences from users. To do so, we collect 300,000 user generated captions and audio pairs along with pairwise preferences through a web interface. We leverage this large-scale feedback to train a reward model and improve our model through RLHF, again consistently outperforming the baseline. Lastly, we combine all reward signals to produce the highest performing model. Additional analysis indicates that the signal extracted from user preferences contains information beyond text adherence and audio quality. This highlights the subjective and complex nature of musical appeal, emphasizing the value of integrating user feedback when improving music generation models.

\section*{Acknowledgments and Contributions}
\looseness=-1
\textbf{Acknowledgments.}
We thank the AI Test Kitchen team of Google for their contributions in designing and deploying the MusicLM experience to users at scale: Sai Kiran Gorthi, Kristin Yim, Phillip Maier, Sejal Khatri, Shirley Leung, Yilun Liu, Yi Yao and Elias Roman. We thank other members and contributors of MusicLM: Timo Denk, Mauricio Zuluaga, Marco Tagliasacchi, Matt Sharifi, Michael Dooley, Christian Frank and Hema Manickavasagam. We also thank Robert Dadashi, Nando de Freitas and Doug Eck for their valuable feedback on earlier drafts of the paper. Finally, we thank the individuals who designed and built the RL training infrastructure used in this paper: Johan Ferret, Nino Vieillard, Alexis Jacq, Sabela Ramos, Piotr Stanczyk, Danila Sinopalnikov, Amélie Héliou, and Nikola Momchev. We are also grateful to all the users of MusicLM for their valuable feedback and their contribution in creating MusicRL.

\textbf{Contributions.}
Geoffrey Cideron and Andrea Agostinelli (main investigators for the project), Neil Zeghidour (core contributions on initial design, supervision, quality reward modelling, paper writing), Sertan Girgin (core infrastructure), Mauro Verzetti (infrastructure, core quality reward modelling), Léonard Hussenot (supervision, core paper writing), Victor Ungureanu (core quality reward modelling), Matej Kastelic (user data infrastructure), Zalán Borsos, Brian McWilliams and Damien Vincent (core MusicLM modelling), Olivier Bachem and Olivier Pietquin (supervision, initial design), and Matthieu Geist (supervision).

\bibliographystyle{abbrvnat}
\nobibliography*
\bibliography{main}

\newpage
\appendix
\onecolumn
\section{Qualitative Evaluation.}
\label{qualitative-evaluation}
For the qualitative evaluation, the list of the 101 diverse prompts is the following:

\noindent\fbox{%
    \parbox{0.9\columnwidth}{\small%
'A mid-tempo country chorus with a choir counter melody ',

 'grunge with a drum n bass beat',
 
 'An eerie, powerful slow metal guitar riff with drums backing that builds tension and anticipation.',
 
 'A wistful, nostalgic indie folk-pop song with a strong bass and a deep male voice',
 
 'Reggeaton with deep bass and a rapping voice',
 
 'A modern, romantic slow waltz played by a jazz trio',
 
 'A rock-steady intro with trumpets providing the backing to a gentle guitar',
 
 'a funky disco song with a bass player',
 
 'A slow rumba composition with a female voice supported by a piano a percussions',
 
 'A sad pop melody with piano and strings accompaniment',
 
 'Chinese music instruments in futuristic theme, fast pace',
 
 'A frantic drum machine beat and pew-pew laser noises fill the cavernous warehouse rave',
 
 'fast, classical music with a church organ with an eerie feeling, for a dark thriller soundtrack',
 
 'A fast, energetic tango played by an accordion and a violin',
 
 'A mellow british-rock acoustic chorus',
 
 'Repetitive house music with strong percussive line',
 
 "The sitar's slow, meandering melody was accompanied by the tabla's steady beat, creating a sound that was both calming and enchanting.",
 
 'Energetic punk rock with a female voice singing',
 
 'a cheerful children song with a simple xylophone backing',
 
 'An energetic gospel choir performance',
 
 'slow, mellow, and instrumental new age music for meditation.',
 
 'Flamenco performance full of energy',
 
 'Melodic danceable brazilian music with percussions.',
 
 'An indie-rock chorus is played by a male singer with a small band backing.',
 
 'epic movie soundtrack',
 
 "The K-pop group's powerful vocals were accompanied by a lush string arrangement, creating a truly epic soundscape.",
 
 'A funk bass intro with a guitar playing short chords and a drums backing',
 
 'Salsa music played by an orchestra',
 
 'A small band plays a latin danceable song',
 
 'A whistling tune for a western duel soundtrack',
 
 'A samba beat and a lively chorus combine to create a festive atmosphere.',
 
 'A jazzy pop song played by a big band',
 
 'a ska-punk trumpet riff supported by an up-beat guitar',
 
 'male bass low grave voice male-singing a medieval song with a mandolin',
 
 'a fast symphonic metal guitar solo with a choir backing',
 
 'chorus of a sweet acoustic rock ballad',
 
 'A bluesy piano riff drives the band as they belt out a soulful tune.',
 
 'A slow, swing pop song with piano and drums backing',
 
 'A fusion of reggaeton and electronic dance music, with a spacey, otherworldly sound.',
 
 'A marching band plays a catchy tune',
 
 'A classical orchestral waltz for a costume dance',
 
 'Irish folk chorus with a mandolin and team whistle',
 
 'A male voice sings a pop anthem accompanied by his piano',
 
 'A catchy pop tune is sung on top a dance drumbeat',
 
 "The soprano's voice soared over the delicate accompaniment of the piano, filling the opera house with beauty and emotion.",
 
 'Rap song with a female melodic line',

 'a reggae song with guitar and singing',
 
 'A corny pop chorus sung by a female voice with a lot of autotune',

    }%
}

\noindent\fbox{%
    \parbox{0.9\textwidth}{\small%
 
 "The marimba's soulful melody was accompanied by the steady beat of the drums, creating a bluesy sound that was both melancholy and uplifting.",
 
 'A gospel choir sings on top a metal guitar backing',
 
 'A powerful female voice sings with soul and energy over a driving drum beat.',
 
 'A repetitive lullaby sung by a female voice with a carillon backing',
 
 'Traditional fast song played by a male voice with an accordion backing',
 
 'An up-beat reggae with a deep male voice and a piano striking the chords',
 
 'Slow, melodic music backed by a sitar and strings.',
 
 'Funky piece with a strong, danceable beat, a prominent bassline and a keyboard melody.',
 
 "A danceable, fast and cheerful swing tune from the 50's",
 
 'a professional solo cellist playing a sad melody for solo cello on the cello, high quality recording',
 
 'A rock guitar riff, a slide guitar solo and a flute melody create a lively, upbeat sound.',
 
 'an a cappella chorus singing a christmas song',
 
 'nice ragtime guitar chord progression',
 
 "A cheerful R'n'B song is played by two singers with a trumpet melody",
 
 'A dance song with a fast melody taken from sampled voice, giving the impression of percussions',
 
 'a gospel song with a female lead singer',
 
 'a nostalgic tune played by accordion band',
 
 'A mariachi song with an epic twist and symphonic orchestra backing',
 
 'A middle-easter tune with percussions and flutes',
 
 'Jazz composition for piano and trumpet',
 
 'A slow blues intro with a harmonica and minimal backing.',
 
 'The experimental modular synthesizer created a unique soundscape by combining the sounds of water with electronic music.',
 
 'a cheerful ragtime with guitar',
 
 'Industrial techno sounds, with hypnotic rhythms. Strings playing a repetitive melody creates an unsettling atmosphere.',
 
 'The microphone picked up the soulful, funky scream of the lead singer as he reached the climax of the song.',
 
 'The snare drum and lute played a lively duet, with the snare drum providing a steady beat and the lute playing a melody on top.',
 
 'The two rappers traded verses over a pulsating synth beat, creating a sound that was both energetic and infectious.',
 
 'A bagpipe is playing an aggressive tune with a punk backing',
 
 'A string quartet plays a lively tune.',
 
 'A very fast piano cadenza that is hard to play.',
 
 'A lone harmonica plays a haunting melody over the sound of the wind blowing through the desert.',
 
 'An aggressive, but sad punk verse, with a prominent slow guitar melody and dark bass line.',
 
 'a band playing cumbia in a boat along the magdalena river in colombia',
 
 'A slow jamaican ska song with an organ backing',
 
 'The gramophone needle crackled and hissed as it spun across the vinyl record, filling the room with a warm, nostalgic sound.',
 
 'fast piano toccata',
 
 "Romantic R'n'B song with a warm female voice",
 
 'A cheerful bollywood-style group dance',
 
 'Dance music with a melodic synth line and arpeggiation',
 
 'The wooden bongo drums beat a deep, resonating bass as the dancers move their bodies to the music.',
 
 'a tenor singing with a backing guitar',
 
 'Slow trap song with a lot of reverb and autotune',
 
 'A syncopated progressive rock tune with a saxophone ',
 
 'A syncopated drum beat backs a hard rock guitar riff',
 
 'a gregorian chant',
 
 'A danceable folk waltz is played by an accordion',
 
 'A bagpipe is playing a fast catchy tune in a dance-pop song',

    }%
}

\noindent\fbox{%
    \parbox{0.9\textwidth}{\small%

 'A full orchestra playing a violin concerto from the 1800s',
 
 'The trap beat was layered with acoustic string sounds creating a catchy chorus.',
 
 'A church choir sings a high-pitched, soothing melody.',
 
 'An energetic dance-pop song, sang by a powerful female voice',
 
 'An harmonica plays a melancholic solo over an acoustic guitar',
 
 'A fast rap chorus is sung on top of a simple catchy tune'
}
}

\section{Additional Quantitative Evaluation Plots}
\label{curves-vs-step}
\begin{figure*}[ht!]
\centering
\begin{subfigure}{}
    \includegraphics[width=0.48\textwidth]{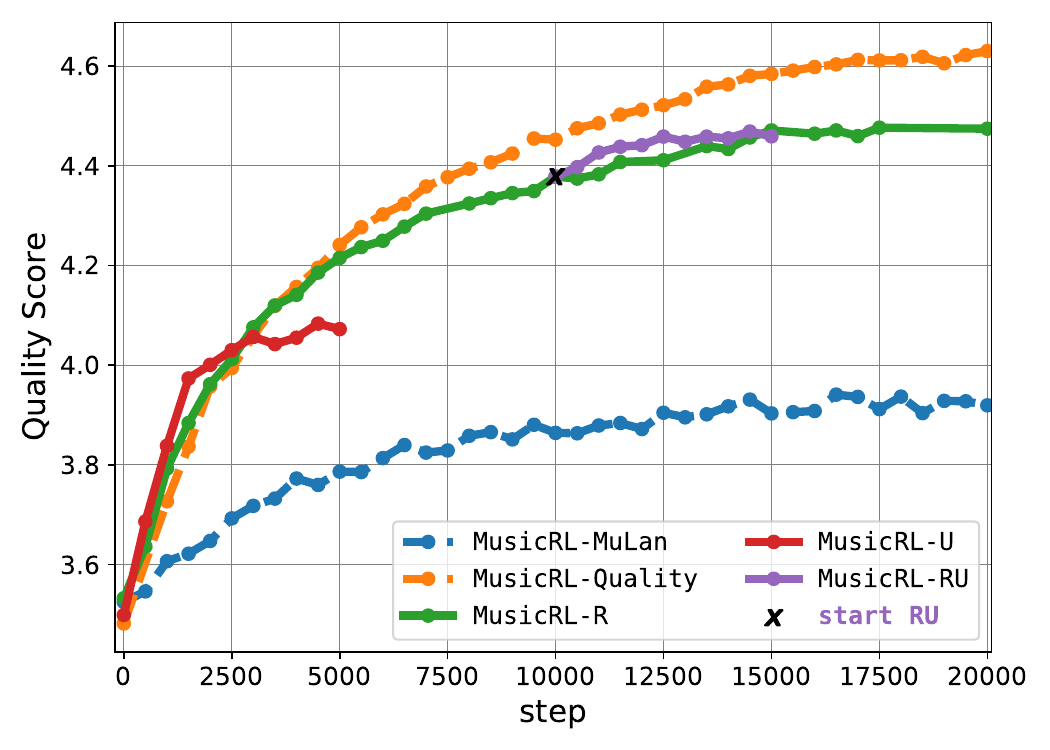}
    \label{fig:mosic-step}
\end{subfigure}
\hfill
\begin{subfigure}{}
    \includegraphics[width=0.48\textwidth]{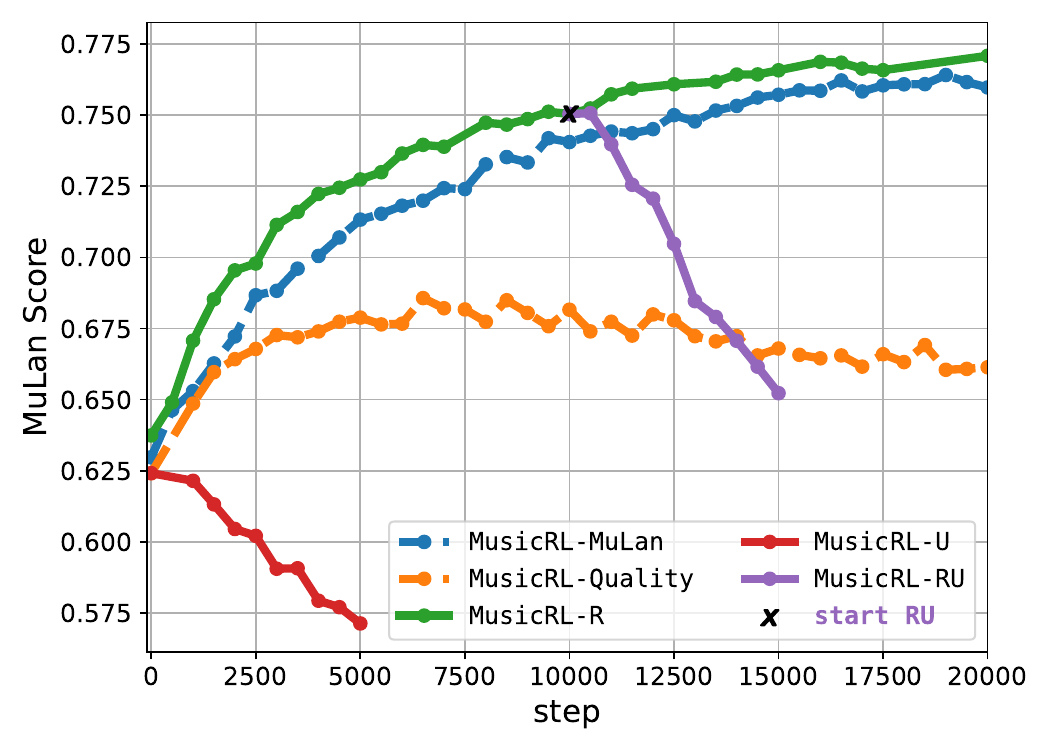}
    \label{fig:mulan-step}
\end{subfigure}
\hfill
\caption{Quality (left) or MuLan score (right) vs step for the RL-finetuned models. The black cross corresponds to the checkpoint used to start the training of \musicrlhfru{}. RL-finetuning successfully optimises the quality and the MuLan scores (\musicrlhfr{}). Additionally, optimizing the user preference reward (\musicrlhfru{}, \musicrlhfru{}) improves the quality score while the MuLan score starts to be significantly impacted when the model over optimise the user preference reward.}
\label{mosic-mulan-vs-step}
\end{figure*}

\begin{figure}[ht!]
\centering
\includegraphics[width=.48\textwidth]{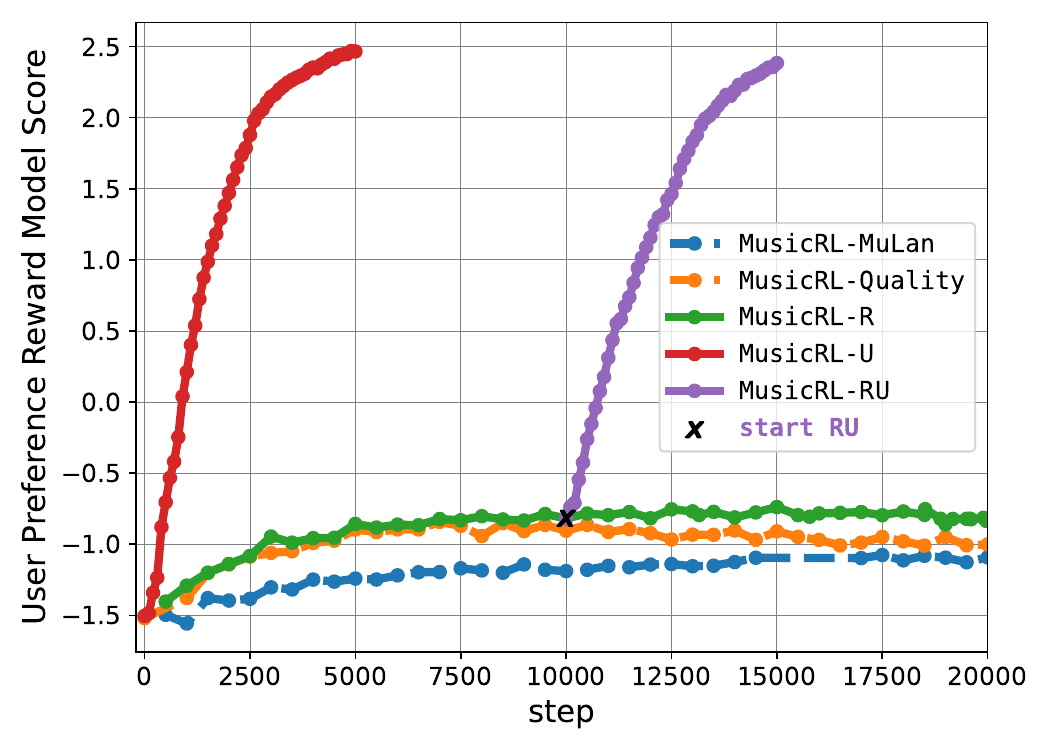}
\caption{User Preference Reward Model Score for the different RL-finetuned models. The black cross corresponds to the checkpoint used to start the training of \musicrlhfru{}. RL-finetuning successfully improves the user preference reward model score of the generations (see \musicrlhfu{} and \musicrlhfru{} curves). When trained on other rewards (MuLan and/or quality) the user preference reward model score slightly improves.}
\label{rm-vs-step}
\end{figure}

Figure~\ref{mosic-mulan-vs-step} and Figure~\ref{rm-vs-step} show the progress of the RL-finetuned models along training as measured by the three reward signals.

\section{Advantages of User Data}
\label{r-to-u}
In Section~\ref{results}, we show that we could leverage a model that was trained with human rater data to improve a music generation model with RL. However, rater data have some limitations and biases. 

In behavioural sciences, the \textit{ecological validity}\footnote{https://en.wikipedia.org/wiki/Ecological\_validity} of a lab study refers its potential of generalization to the real world \citep{conceptvalidity}. In the context of music, it is crucial to experiment on real-world settings \citep{musicvalidity}. \citet{ecologicalgap} explore the \textit{ecological validity} concept in the context of interface design and say that "User-related ecological gaps are caused by characteristics of users - such as what motivates them, their cognitive abilities, preferences, and habits - that may vary between the lab and the target environment." The concept of user-related ecological gaps is particularly relevant for the finetuning and the evaluation of large language models as the raters and users are often dissimilar. 

\textbf{Population Gap.} Raters are often not representative of the user population especially as the rating task is often outsourced to crowdsourcing services which employ people in different countries than the ones the model is deployed in e.g. Amazon Mechanical Turk\footnote{https://www.mturk.com/} proposes a global workforce for rating tasks. This population difference creates biases such as cultural biases which can impact the music preferences \citep{crosscultural}. 

\textbf{Motivation Gap.} As mentioned in \citet{ecologicalgap}, the \textit{motivation gap} which corresponds to the difference of motivations between the different users can have a significant effect on the results. In our context, while the users of music generation models have a genuine interest in playing with the model, the incentive of the raters are very different. Hence, for rating tasks, it is crucial to give specific set of instructions to make sure the raters make their decisions aligned with what the creator of the rating task would expect which also can be a source of biases. Whereas for users, we are interested in general interactions where no instructions are given. 

\textbf{Dataset Size.} Due to the cost of rater data, the number of collected human preference is often below 100,000 \citep{ziegler2019fine,imagerlhf,stiennon2020learning}. On the other hand, the number of user interactions can be orders of magnitude higher once a model is deployed.